\documentclass[11pt,a4paper]{article}
\usepackage[hyperref]{acl2019}
\usepackage{times}
\usepackage{latexsym}
\usepackage{graphicx}
\usepackage{multirow}
\usepackage{url}
\usepackage{amssymb}
\usepackage{amsmath}
\usepackage{enumitem}
\usepackage{url}
\usepackage{color,soul}
\usepackage{subfig}

\aclfinalcopy 

\title{Transferable Multi-Domain State Generator for Task-Oriented \\ Dialogue Systems}

\author{Chien-Sheng Wu$^\dag$\thanks{ { } Work partially done while the first author was an intern at Salesforce Research.}, Andrea Madotto$^\dag$, Ehsan Hosseini-Asl$^\ddag$, Caiming Xiong$^\ddag$,\\      
\textbf{Richard Socher$^\ddag$ and Pascale Fung$^\dag$} \\
    $^\dag$The Hong Kong University of Science and Technology \\
    $^\ddag$Salesforce Research \\
    \texttt{jason.wu@connect.ust.hk}}

\date{}

\begin{document}
\maketitle
\begin{abstract}
Over-dependence on domain ontology and lack of knowledge sharing across domains are two practical and yet less studied problems of dialogue state tracking. 
Existing approaches generally fall short in tracking unknown slot values during inference and often have difficulties in adapting to new domains.
In this paper, we propose a
\textbf{TRA}nsferable \textbf{D}ialogue stat\textbf{E} generator (\textbf{TRADE}) 
that generates dialogue states from utterances using a copy mechanism, facilitating knowledge transfer when predicting \textit{(domain, slot, value)} triplets not encountered during training.
Our model is composed of an utterance encoder, a slot gate, and a state generator, which are shared across domains. 
Empirical results demonstrate that TRADE achieves state-of-the-art joint goal accuracy of 48.62\% for the five domains of MultiWOZ, a human-human dialogue dataset.
In addition, we show its transferring ability by simulating zero-shot and few-shot dialogue state tracking for unseen domains.
TRADE achieves 60.58\% joint goal accuracy in one of the zero-shot domains, and is able to adapt to few-shot cases without forgetting already trained domains. 

\end{abstract}

\section{Introduction}
\label{sec:INTRO}
\begin{figure}[t]
\centering
\includegraphics[width=\linewidth]{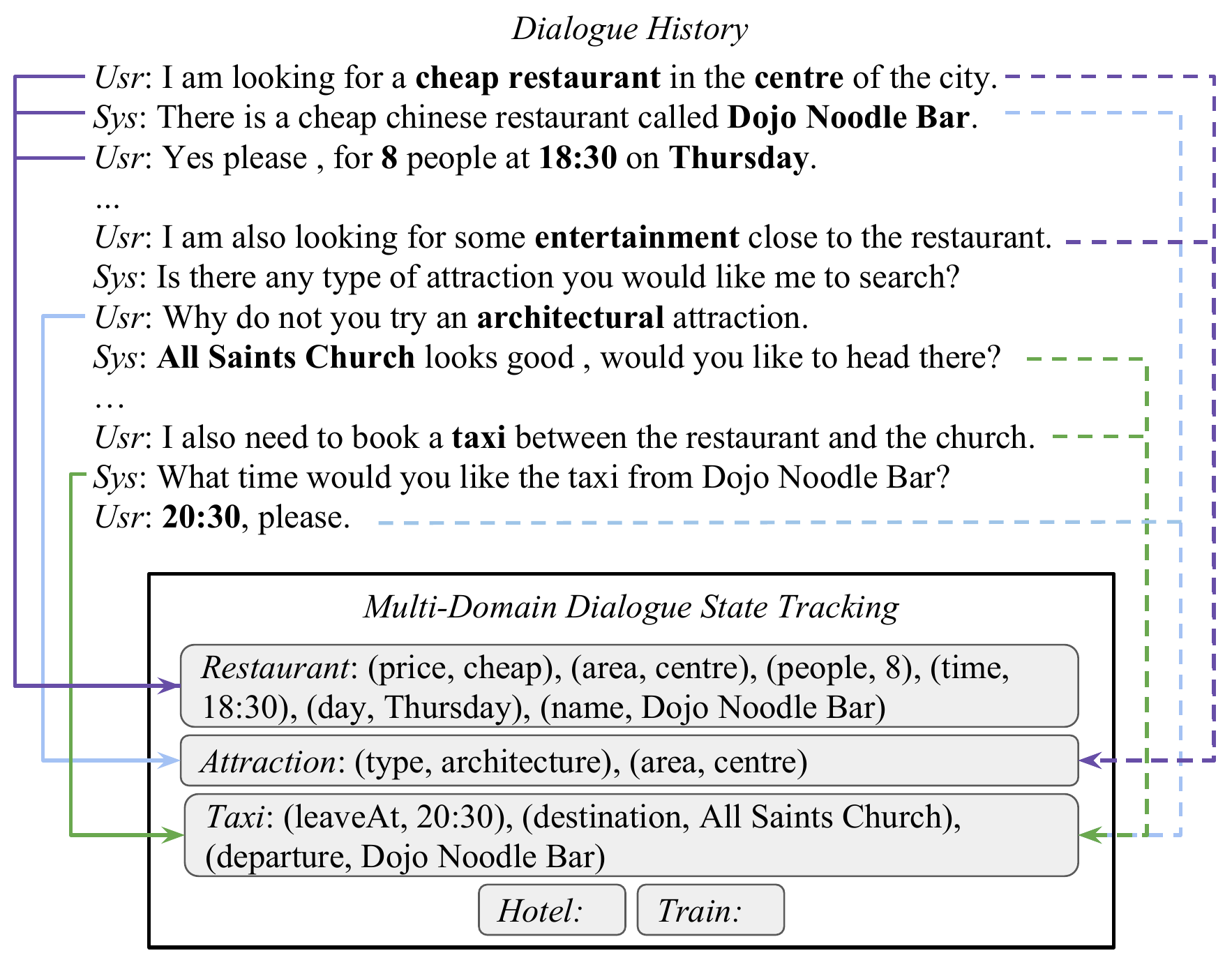}
\caption{An example of multi-domain dialogue state tracking in a conversation. The solid arrows on the left are the single-turn mapping, and the dot arrows on the right are multi-turn mapping. The state tracker needs to track slot values mentioned by the user for all the slots in all the domains.}
\label{example}
\end{figure}

Dialogue state tracking (DST) is a core component in task-oriented dialogue systems, such as restaurant reservation or ticket booking.
The goal of DST is to extract user goals/intentions expressed during conversation and to encode them as a compact set of dialogue states, i.e., a set of slots and their corresponding values.
For example, as shown in Fig.~\ref{example}, \textit{(slot, value)} pairs such as \textit{(price, cheap)} and \textit{(area, centre)} are extracted from the conversation.
Accurate DST performance is crucial for appropriate dialogue management, where user intention determines the next system action and/or the content to query from the databases.

Traditionally, state tracking approaches are based on the assumption that ontology is defined in advance, where all slots and their values are known.
Having a predefined ontology can simplify DST into a classification problem and improve performance ~\cite{henderson2014word, NBT, GLAD}. 
However, there are two major drawbacks to this approach: 
1) A full ontology is hard to obtain in advance~\cite{P18-1134PtrNet}.
In the industry, databases are usually exposed through an external API only, which is owned and maintained by others. It is not feasible to gain access to enumerate all the possible values for each slot.
2) Even if a full ontology exists, the number of possible slot values could be large and variable. 
For example, a restaurant name or a train departure time can contain a large number of possible values. Therefore, many of the previous works that are based on neural classification models may not be applicable in real scenario.

~\citet{multiwoz} recently introduced a multi-domain dialogue dataset (MultiWOZ), which adds new challenges in DST due to its mixed-domain conversations.
As shown in Fig.~\ref{example}, a user can start a conversation by asking to reserve a restaurant, then requests information regarding an attraction nearby, and finally asks to book a taxi. 
In this case, the DST model has to determine the corresponding domain, slot and value at each turn of dialogue, which contains a large number of combinations in the ontology, i.e., 30 (domain, slot) pairs and over 4,500 possible slot values in total.
Another challenge in the multi-domain setting comes from the need to perform multi-turn mapping.
Single-turn mapping refers to the scenario where the \textit{(domain, slot, value)} triplet can be inferred from a single turn, while in multi-turn mapping, it should be inferred from multiple turns which happen in different domains.
For instance, the \textit{(area, centre)} pair from the \textit{attraction} domain in Fig.~\ref{example} can be predicted from the \textit{area} information in the \textit{restaurant} domain, which is mentioned in the preceding turns.

To tackle these challenges, we emphasize that DST models should share tracking knowledge across domains. There are many slots among different domains that share all or some of their values.
For example, the \textit{area} slot can exist in many domains, e.g., \textit{restaurant}, \textit{attraction}, and \textit{taxi}. Moreover, the~\textit{name} slot in the \textit{restaurant} domain can share the same value with the \textit{departure} slot in the \textit{taxi} domain.
Additionally, to enable the DST model to track slots in unseen domains, transferring knowledge across multiple domains is imperative. 
We expect DST models can learn to track some slots in zero-shot domains by learning to track the same slots in other domains.

In this paper, we propose a transferable dialogue state generator (TRADE) for multi-domain task-oriented dialogue state tracking. 
The simplicity of our approach and the boost of the performance is the main advantage of TRADE.
Contributions in this work are summarized as~\footnote{The code is released at \url{github.com/jasonwu0731/trade-dst}}:
\begin{itemize}[leftmargin=*]
    \item To overcome the multi-turn mapping problem, TRADE leverages its context-enhanced slot gate and copy mechanism to properly track slot values mentioned anywhere in dialogue history.
  
  \item By sharing its parameters across domains, and without requiring a predefined ontology, TRADE can share knowledge between domains to track unseen slot values, achieving state-of-the-art performance on multi-domain DST.

  \item TRADE enables zero-shot DST by leveraging the domains it has already seen during training. If a few training samples from unseen domains are available, TRADE can adapt to new few-shot domains without forgetting the previous domains.
  
\end{itemize}

\begin{figure*}[t]
\centering
\includegraphics[width=\linewidth]{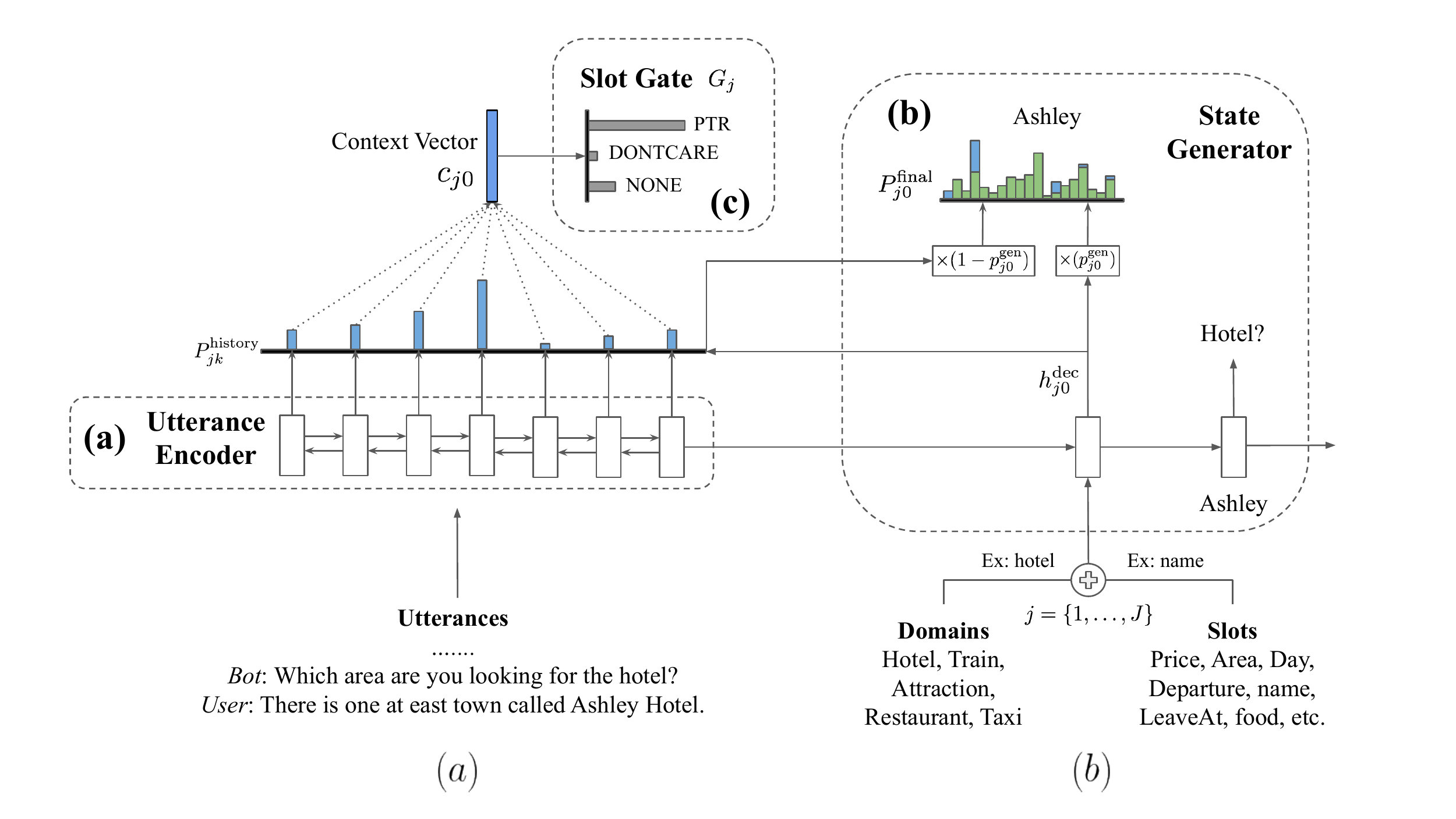}
\caption{The architecture of the proposed TRADE model, which includes (a) an utterance encoder, (b) a state generator, and (c) a slot gate, all of which are shared among domains. The state generator will decode $J$ times independently for all the possible \textit{(domain, slot)} pairs. At the first decoding step, state generator will take the $j$-th \textit{(domain, slot)} embeddings as input to generate its corresponding slot values and slot gate. The slot gate predicts whether the $j$-th \textit{(domain, slot)} pair is triggered by the dialogue.}
\label{model}
\end{figure*}

\section{TRADE Model}
The proposed model in Fig.~\ref{model} comprises three components: an utterance encoder, a slot gate, and a state generator. 
Instead of predicting the probability of every predefined ontology term, our model directly generates slot values.
Similar to~\citet{Q17-1024} for multilingual neural machine translation, we share all the model parameters, and the state generator starts with a different start-of-sentence token for each \textit{(domain, slot)} pair.

The utterance encoder encodes dialogue utterances into a sequence of fixed-length vectors.
To determine whether any of the \textit{(domain, slot)} pairs are mentioned, the context-enhanced slot gate is used with the state generator. 
The state generator decodes multiple output tokens for all \textit{(domain, slot)} pairs independently to predict their corresponding values.
The context-enhanced slot gate predicts whether each of the pairs is actually triggered by the dialogue via a three-way classifier. 

Let us define $X = \{(U_1, R_1), \dots, (U_T, R_T)\}$ as the set of user utterance and system response pairs in $T$ turns of dialogue,
and $B = \{B_1,\dots,B_T\}$ as the dialogue states for each turn. 
Each $B_t$ is a tuple (domain:$D_n$, slot:$S_m$, value:$Y^{\text{value}}_j$), where $D = \{D_1, \dots, D_N\}$ are the $N$ different domains, and $S=\{S_1,\dots,S_M\}$ are the $M$ different slots.
Assume that there are $J$ possible \textit{(domain, slot)} pairs, and $Y^{\text{value}}_j$ is the true word sequence for $j$-th \textit{(domain ,slot)} pair.

\subsection{Utterance Encoder}
Note that the utterance encoder can be any existing encoding model.
We use bi-directional gated recurrent units (GRU) ~\cite{chung2014empirical} to encode the dialogue history.
The input to the utterance encoder is denoted as \textit{history} $X_{t} = [U_{t-l},R_{t-l},\dots,U_t,R_t]\in \mathbb{R}^{|X_{t}| \times d_{emb}}$, which is the concatenation of all words in the dialogue history. 
$l$ is the number of selected dialogue turns and $d_{emb}$ indicates the embedding size. 
The encoded dialogue history is represented as $H_{t} = [h_1^{\text{\text{enc}}},\dots,h_{|X_{t}|}^{\text{enc}}] \in \mathbb{R}^{|X_{t}| \times d_{hdd}}$, where $d_{hdd}$ is the hidden size.
As mentioned in Section~\ref{sec:INTRO}, due to the multi-turn mapping problem, the model should infer the states across a sequence of turns. Therefore, we use the recent dialogue history of length $l$ as the utterance encoder input, rather than the current utterance only. 

\subsection{State Generator}
To generate slot values using text from the input source, a copy mechanism is required.
There are three common ways to perform copying, i.e., index-based copy~\cite{vinyals2015pointer}, hard-gated copy ~\cite{gulcehre2016pointing, madotto2018mem2seq,wu2019global} and soft-gated copy ~\cite{see2017PG, mccann2018natural}.
The index-based mechanism is not suitable for DST task because the exact word(s) of the true slot value are not always found in the utterance.
The hard-gate copy mechanism usually needs additional supervision on the gating function. 
As such, we employ soft-gated pointer-generator copying to combine a distribution over the vocabulary and a distribution over the dialogue history into a single output distribution.

We use a GRU as the decoder of the state generator to predict the value for each \textit{(domain, slot)} pair, as shown in Fig.~\ref{model}.
The state generator decodes $J$ pairs independently. We simply supply the summed embedding of the domain and slot as the first input to the decoder.
At decoding step $k$ for the $j$-th \textit{(domain, slot)} pair, the generator GRU takes a word embedding $w_{jk}$ as its input and returns a hidden state $h^{\text{\text{dec}}}_{jk}$. 
The state generator first maps the hidden state $h^{\text{dec}}_{jk}$ into the vocabulary space $P^{\text{vocab}}_{jk}$ using the trainable embedding $E \in \mathbb{R}^{|V| \times d_{hdd}}$, where $|V|$ is the vocabulary size. 
At the same time, the $h^{\text{dec}}_{jk}$ is used to compute the history attention $P^{\text{history}}_{jk}$ over the encoded dialogue history $H_{t}$: 
\begin{equation}
\resizebox{0.85\linewidth}{!}{$
    \begin{array}{c}
        P^{\text{vocab}}_{jk}= \text{Softmax}(E\cdot  (h^{\text{dec}}_{jk})^\top) \in \mathbb{R}^{|V|}, \\
        P^{\text{\text{history}}}_{jk} = \text{Softmax}(H_{t}\cdot  (h^{\text{dec}}_{jk})^\top) \in \mathbb{R}^{|X_t|}.
    \end{array}
$}
\label{prob}
\end{equation}
The final output distribution $P^{\text{final}}_{jk}$ is the weighted-sum of two distributions,
\begin{align}
\begin{split}
        P^{\text{final}}_{jk} 
        &= p^{\text{gen}}_{jk} \times P^{\text{vocab}}_{jk} \\
        &+  (1-p^{\text{gen}}_{jk}) \times P^{\text{\text{history}}}_{jk} \in \mathbb{R}^{|V|}.
    \label{pg_mix}
\end{split}
\end{align}
The scalar $p^{\text{gen}}_{jk}$ is trainable to combine the two distributions, which is computed by
\begin{equation}
\resizebox{0.88\linewidth}{!}{$
    \begin{array}{c}
    p^{\text{gen}}_{jk} = \text{Sigmoid}(W_1\cdot  [h^{\text{dec}}_{jk}; w_{jk}; c_{jk}]) \in \mathbb{R}^{1}, \\
    c_{jk} = P^{\text{\text{history}}}_{jk}\cdot H_{t} \in \mathbb{R}^{d_{hdd}}
    
    \end{array}
\label{p_gen}
$}
\end{equation}
where $W_1$ is a trainable matrix and $c_{jk}$ is the context vector. 
Note that due to Eq (\ref{pg_mix}), our model is able to generate words even if they are not pre-defined in the vocabulary.

\subsection{Slot Gate}
Unlike single-domain DST problems, where only a few slots that need to be tracked, e.g., four slots in WOZ~\cite{Wen-WOZ}, and eight slots in DSTC2~\cite{henderson2014dstc2}, there are a large number of \textit{(domain, slot)} pairs in multi-domain DST problems. 
Therefore, the ability to predict the domain and slot at current turn $t$ becomes more challenging.

Our context-enhanced slot gate $G$ is a simple three-way classifier that maps a context vector taken from the encoder hidden states $H_{t}$ to a probability distribution over \textit{ptr}, \textit{none}, and \textit{dontcare} classes.
For each \textit{(domain, slot)} pair, if the slot gate predicts \textit{none} or \textit{dontcare}, we ignore the values generated by the decoder and fill the pair as ``not-mentioned'' or ``does not care''.
Otherwise, we take the generated words from our state generator as its value.
With a linear layer parameterized by $W_g\in \mathbb{R}^{3 \times d_{hdd}}$, the slot gate for the $j$-th \textit{(domain, slot)} pair is defined as 
\begin{equation}
    G_j = \text{Softmax}(W_g\cdot  (c_{j0})^\top) \in \mathbb{R}^{3},
\label{gate}
\end{equation}
where $c_{j0}$ is the context vector computed in Eq (\ref{p_gen}) using the first decoder hidden state.

\subsection{Optimization}
During training, we optimize for both the slot gate and the state generator. 
For the former, the cross-entropy loss $L_{g}$ is computed between the predicted slot gate $G_{j}$ and the true one-hot label $y^{\text{gate}}_j$, 
\begin{equation}
    L_g = \sum_{j=1}^{J} - \log(G_j\cdot  (y^{\text{gate}}_j)^\top).
\end{equation}
For the latter, another cross-entropy loss $L_v$ between $P^{\text{final}}_{jk}$ and the true words $Y_{j}^{\text{label}}$ is used. We define $L_v$ as 
\begin{equation}
L_v = \sum_{j=1}^{J} \sum_{k=1}^{|Y_j|} - \log(P^{\text{final}}_{jk}\cdot (y^{\text{value}}_{jk})^\top).
\end{equation}
$L_v$ is the sum of losses from all the \textit{(domain, slot)} pairs and their decoding time steps. We optimize the weighted-sum of these two loss functions using hyper-parameters $\alpha$ and $\beta$,
\begin{equation}
\label{eq:loss}
L = \alpha L_g + \beta L_v . 
\end{equation}

\section{Unseen Domain DST}
In this section, we focus on the ability of TRADE to generalize to an unseen domain by considering zero-shot transferring and few-shot domain expanding.
In the zero-shot setting, we assume we have no training data in the new domain, while in the few-shot case, we assume just 1\% of the original training data in the unseen domain is available (around 20 to 30 dialogues). 
One of the motivations to perform unseen domain DST is because collecting a large-scale task-oriented dataset for a new domain is expensive and time-consuming ~\cite{multiwoz}, and there are a large amount of domains in realistic scenarios.

\subsection{Zero-shot DST}
Ideally, based on the slots already learned, a DST model is able to directly track those slots that are present in a new domain. 
For example, if the model is able to track the \textit{departure} slot in the \textit{train} domain, then that ability may transfer to the \textit{taxi} domain, which uses similar slots.
Note that generative DST models take the dialogue context/history $X$, the domain $D$, and the slot $S$ as input and then generate the corresponding values $Y^{\text{value}}$.
Let $(X, D_{\text{source}}, S_{\text{source}},Y^{\text{value}}_{\text{source}})$ be the set of samples seen during the training phase and $(X, D_{\text{\text{target}}}, S_{\text{target}},Y^{\text{value}}_{\text{target}})$ the samples which the model was not trained to track. 
A zero-shot DST model should be able to generate the correct values of $Y^{\text{value}}_{\text{target}}$ given the context $X$, domain $D_{\text{target}}$, and slot $S_{\text{target}}$, without using any training samples. 
The same context $X$ may appear in both source and target domains but the pairs ($D_{\text{target}}, S_{\text{target}}$) are unseen. 
This setting is extremely challenging if no slot in $S_{\text{target}}$ appears in $S_{\text{source}}$, since the model has never been trained to track such a slot.

\subsection{Expanding DST for Few-shot Domain}
\label{sec:cl}
In this section, we assume that only a small number of samples from the new domain $(X, D_{\text{target}}, S_{\text{target}}, Y^{\text{value}}_{\text{target}})$ are available, and the purpose is to evaluate the ability of our DST model to transfer its learned knowledge to the new domain without forgetting previously learned domains.
There are two advantages to performing few-shot domain expansion: 
1) being able to quickly adapt to new domains and obtain decent performance with only a small amount of training data;
2) not requiring retraining with all the data from previously learned domains, since the data may no longer be available and retraining is often very time-consuming.

Firstly, we consider a straightforward naive baseline, i.e., fine-tuning with no constraints. Then, we employ two specific continual learning techniques: elastic weight consolidation (EWC)~\cite{kirkpatrick2017overcoming} and gradient episodic memory (GEM)~\cite{lopez2017gradient} to fine-tune our model.
We define $\Theta_{S}$ as the model's parameters trained in the source domain, and $\Theta$ indicates the current optimized parameters according to the target domain data. 

EWC uses the diagonal of the Fisher information matrix $F$ as a regularizer for adapting to the target domain data. This matrix is approximated using samples from the source domain.
The EWC loss is defined as
\begin{equation}
    L_{ewc}(\Theta) = L(\Theta) + \sum_i \frac{\lambda}{2} F_i (\Theta_i - \Theta_{S,i})^2,
\end{equation}
where $\lambda$ is a hyper-parameter.
Different from EWC, GEM keeps a small number of samples $K$ from the source domains, and, while the model learns the new target domain, a constraint is applied on the gradient to prevent the loss on the stored samples from increasing. 
The training process is defined as: 
\begin{align}
\begin{split}
    &\text{Minimize}_{\Theta} \ L(\Theta) \\
    & \text{Subject to} \  L(\Theta, K) \leq L(\Theta_S, K), 
\label{contrain}
\end{split}
\end{align}
where $L(\Theta, K)$ is the loss value of the $K$ stored samples. 
\citet{lopez2017gradient} show how to solve the optimization problem in Eq (\ref{contrain}) with quadratic programming if the loss of the stored samples increases.

\section{Experiments}
\subsection{Dataset}
Multi-domain Wizard-of-Oz \cite{multiwoz} (MultiWOZ) is the largest existing human-human conversational corpus spanning over seven domains, containing 8438 multi-turn dialogues, with each dialogue averaging 13.68 turns.
Different from existing standard datasets like WOZ~\cite{Wen-WOZ} and DSTC2~\cite{henderson2014dstc2}, which contain less than 10 slots and only a few hundred values, MultiWOZ has 30 \textit{(domain, slot)} pairs and over 4,500 possible values.
We use the DST labels from the original training, validation and testing dataset.
Only five domains (\textit{restaurant}, \textit{hotel}, \textit{attraction}, \textit{taxi}, \textit{train}) are used in our experiment because the other two domains (\textit{hospital}, \textit{police}) have very few dialogues (10\% compared to others) and only appear in the training set.
The slots in each domain and the corresponding data size are reported in Table~\ref{DATASET-TABLE}.

\begin{table}[t!]
\begin{center}
\resizebox{\linewidth}{!}{
\begin{tabular}{r|c|c|c|c|c}
\hline
 & \textbf{Hotel} & \textbf{Train} & \textbf{Attraction} & \textbf{Restaurant} & \textbf{Taxi} \\ \hline
\textit{Slots} & \begin{tabular}[c]{@{}c@{}}price,\\ type,\\ parking,\\ stay,\\ day,\\ people,\\ area,\\ stars,\\ internet,\\ name\end{tabular} & \begin{tabular}[c]{@{}c@{}}destination,\\ departure,\\ day,\\ arrive by,\\ leave at,\\ people\end{tabular} & \begin{tabular}[c]{@{}c@{}}area,\\ name,\\ type\end{tabular} & \begin{tabular}[c]{@{}c@{}}food,\\ price,\\ area,\\ name,\\ time,\\ day,\\ people\end{tabular} & \begin{tabular}[c]{@{}c@{}}destination,\\ departure,\\ arrive by,\\ leave by\end{tabular} \\ \hline
\textit{Train} & 3381 & 3103 & 2717 & 3813 & 1654 \\
\textit{Valid} & 416 & 484 & 401 & 438 & 207 \\
\textit{Test} & 394 & 494 & 395 & 437 & 195 \\ \hline
\end{tabular}
}
\end{center}
\caption{The dataset information of MultiWOZ. In total, there are 30 \textit{(domain, slot)} pairs from the selected five domains. The numbers in the last three rows indicate the number of dialogues for train, validation and test sets.}
\label{DATASET-TABLE}
\end{table}

\subsection{Training Details}

\paragraph{Multi-domain Joint Training}
The model is trained end-to-end using the Adam optimizer ~\citep{KingmaB14} with a batch size of 32. 
The learning rate annealing is in the range of $[0.001, 0.0001]$ 
with a dropout ratio of 0.2. Both $\alpha$ and $\beta$ in Eq~(\ref{eq:loss}) are set to one. 
All the embeddings are initialized by concatenating Glove embeddings~\cite{pennington2014glove} and character embeddings ~\cite{hashimoto2016joint}, where the dimension is 400 for each vocabulary word.
A greedy search decoding strategy is used for our state generator since the generated slot values are usually short in length.
In addition, to increase model generalization and simulate an out-of-vocabulary setting, a word dropout is utilized with the utterance encoder by randomly masking a small amount of input tokens, similar to ~\citet{K16-1002}.

\paragraph{Domain Expanding}
For training, we follow the same procedure as in the joint training section, and we run a small grid search for all the methods using the validation set. For EWC, we set different values of $\lambda$ for all the domains, and the optimal value is selected using the validation set. Finally, in GEM, we set the memory sizes $K$ to 1\% of the source domains.

\subsection{Results}
Two evaluation metrics, joint goal accuracy and slot accuracy, are used to evaluate the performance on multi-domain DST. 
The joint goal accuracy compares the predicted dialogue states to the ground truth $B_t$ at each dialogue turn $t$, and the output is considered correct if and only if all the predicted values exactly match the ground truth values in $B_t$.
The slot accuracy, on the other hand, individually compares each (domain, slot, value) triplet to its ground truth label. 

\paragraph{Multi-domain Training}
We make a comparison with the following existing models: MDBT~\cite{MDBT}, GLAD~\cite{GLAD}, GCE~\cite{Nouri2018TowardSN}, and SpanPtr~\cite{P18-1134PtrNet}, and we briefly describe these baselines models below:
\begin{itemize}[leftmargin=*]
  \item MDBT~\footnote{\url{github.com/osmanio2/multi-domain-belief-tracking}}: Multiple bi-LSTMs are used to encode system and user utterances. The semantic similarity between utterances and every predefined ontology term is computed separately. Each ontology term is triggered if the predicted score is greater than a threshold. 
  \item GLAD~\footnote{\url{github.com/salesforce/glad}}: 
  This model uses self-attentive RNNs to learn a global tracker that shares parameters among slots and a local tracker that tracks each slot.
  The model takes previous system actions and the current user utterance as input, and computes semantic similarity with predefined ontology terms.
  
  \item GCE: 
  This is the current state-of-the-art model on the single-domain WOZ dataset~\cite{Wen-WOZ}. It is a simplified and speed up version of GLAD without slot-specific RNNs. 
  
  \item SpanPtr: Most related to our work, this is the first model that applies pointer networks ~\cite{vinyals2015pointer} to the single-domain DST problem, which generates both start and end pointers to perform index-based copying.
\end{itemize}

To have a fair comparison, we modify the original implementation of the MDBT and GLAD models by:
1) adding \textit{name}, \textit{destination}, and \textit{departure} slots for evaluation if they were discarded or replaced by placeholders; and
2) removing the hand-crafted rules of tracking the booking slots such as \textit{stay} and \textit{people} slots if there are any; and
3) creating a full ontology for their model to cover all \textit{(domain, slot, value)} pairs that were not in the original ontology generated by the data provider.

\begin{table}[t!]
\begin{center}
\resizebox{\linewidth}{!}{
\begin{tabular}{r|cc|cc}
\hline
\multicolumn{1}{l|}{} & \multicolumn{2}{c|}{\textbf{MultiWOZ}} & \multicolumn{2}{c}{\textbf{\begin{tabular}[c]{@{}c@{}}MultiWOZ\\ (Only Restaurant)\end{tabular}}} \\ \cline{2-5} 
 & \textit{\textbf{Joint}} & \textit{\textbf{Slot}} & \textit{\textbf{Joint}} & \textit{\textbf{Slot}} \\ \hline
\textit{MDBT} & 15.57 & 89.53 & 17.98 & 54.99 \\ \hline
\textit{GLAD} & 35.57 & 95.44 & 53.23 & 96.54 \\ \hline
\textit{GCE} & 36.27 & 98.42 & 60.93 & 95.85 \\ \hline
\textit{SpanPtr} & 30.28 & 93.85 & 49.12 & 87.89 \\ \hline
\textit{TRADE} & \textbf{48.62} & 96.92 & \textbf{65.35} & 93.28 \\
\hline
\end{tabular}
}
\end{center}
\caption{The multi-domain DST evaluation on MultiWOZ and its single \textit{restaurant} domain. TRADE has the highest joint accuracy, which surpasses current state-of-the-art GCE model. }
\label{JOINT-TABLE}
\end{table}

As shown in Table~\ref{JOINT-TABLE}, TRADE achieves the highest performance, 48.62\% on joint goal accuracy and 96.92\% on slot accuracy, on MultiWOZ.
For comparison with the performance on single-domain, the results on the \textit{restaurant} domain of MultiWOZ are reported as well.
The performance difference between SpanPtr and our model mainly comes from the limitation of index-based copying.
For examples, if the true label for the price range slot is \textit{cheap}, the relevant user utterance describing the restaurant may actually be, for example, \textit{economical}, \textit{inexpensive}, or \textit{cheaply}. 
Note that the MDBT, GLAD, and GCE models each need a predefined domain ontology to perform binary classification for each ontology term, which hinders their DST tracking performance, as mentioned in Section ~\ref{sec:INTRO}. 

\begin{figure}[t]
\centering
\includegraphics[width=0.9\linewidth]{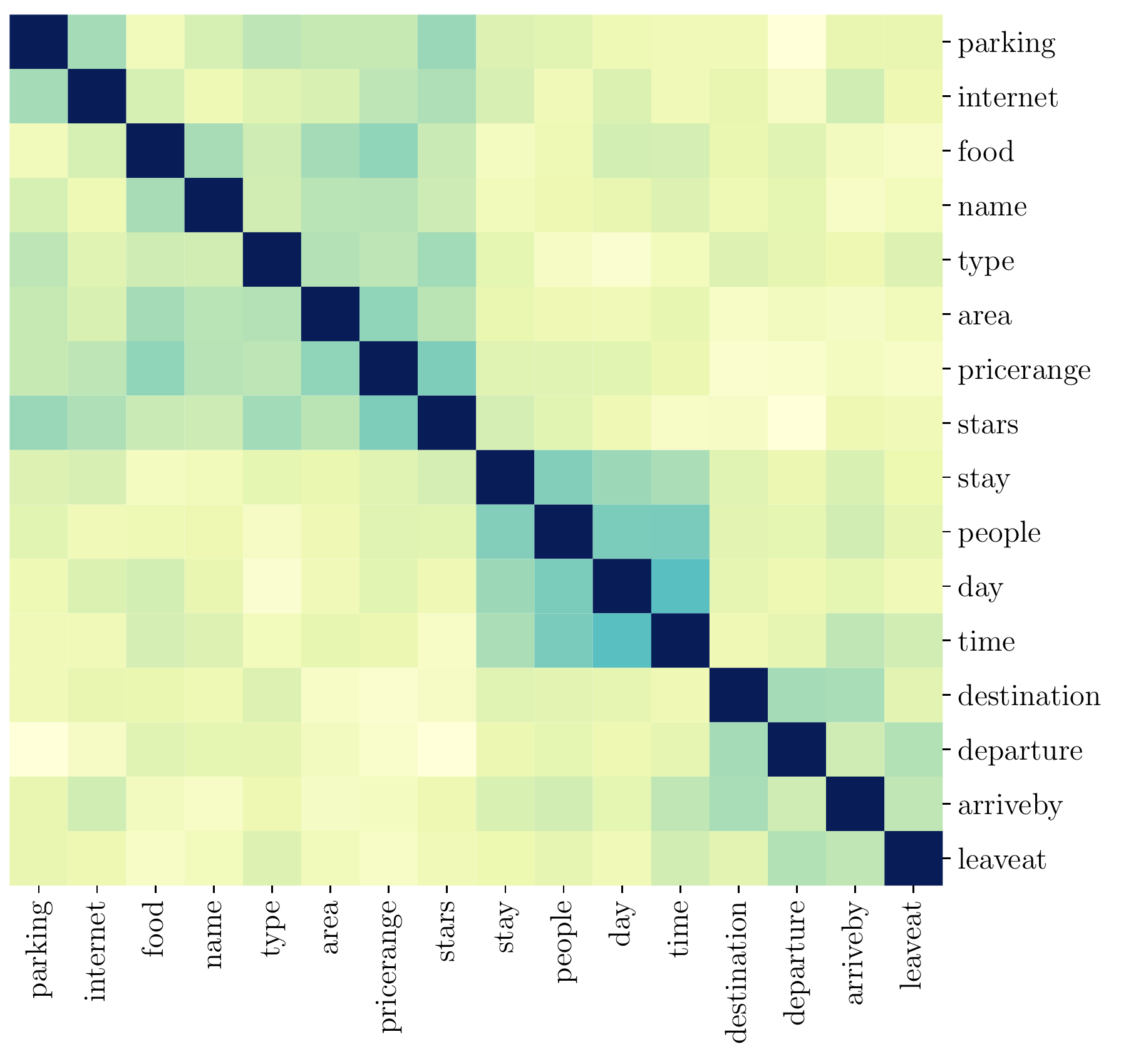}
\caption{Embeddings cosine similarity visualization. The rows and columns are all the possible slots in MultiWOZ. Slots that share similar values or have correlated values learn similar embeddings. For example ~\textit{destination vs. departure} (which share similar values) or ~\textit{price range vs. stars} exhibit high correlation.
}
\label{slot-viz}
\end{figure}

\begin{table*}[t]
\resizebox{\linewidth}{!}{
\begin{tabular}{cr|cc|cc|cc|cc|cc}
\hline
\multicolumn{1}{l}{} &  & \textbf{Joint} & \textbf{Slot} & \textbf{Joint} & \textbf{Slot} & \textbf{Joint} & \textbf{Slot} & \textbf{Joint} & \textbf{Slot} & \textbf{Joint} & \textbf{Slot} \\
\multicolumn{2}{c|}{\textbf{Evaluation on 4 Domains}} & \multicolumn{2}{l|}{\textit{Except Hotel}} & \multicolumn{2}{l|}{\textit{Except Train}} & \multicolumn{2}{l|}{\textit{Except Attraction}} & \multicolumn{2}{l|}{\textit{Except Restaurant}} & \multicolumn{2}{l}{\textit{Except Taxi}} \\ \hline
\multicolumn{2}{c|}{\begin{tabular}[c]{@{}c@{}}Base Model (BM)\\ training on 4 domains\end{tabular}} & 58.98 & 96.75 & 55.26 & 96.76 & 55.02 & 97.03 & 54.69 & 96.64 & 49.87 & 96.77 \\ \hline
\multicolumn{1}{c|}{\multirow{3}{*}{\begin{tabular}[c]{@{}c@{}}Fine-tuning BM\\ on 1\% new domain\end{tabular}}} & \textit{Naive} & 36.08 & 93.48 & 23.25 & 90.32 & 40.05 & 95.54 & 32.85 & 91.69 & 46.10 & 96.34 \\
\multicolumn{1}{c|}{} & \textit{EWC} & 40.82 & 94.16 & 28.02 & 91.49 & 45.37 & 84.94 & 34.45 & 92.53 & \textbf{46.88} & 96.44 \\
\multicolumn{1}{c|}{} & \textit{GEM} & \textbf{53.54} & \textbf{96.27} & \textbf{50.69} & \textbf{96.42} & \textbf{50.51} & \textbf{96.66} & \textbf{45.91} & \textbf{95.58} & 46.43 & \textbf{96.45} \\ 
\hline \hline
\multicolumn{2}{c|}{\textbf{Evaluation on New Domain}} & \multicolumn{2}{c|}{\textit{Hotel}} & \multicolumn{2}{c|}{\textit{Train}} & \multicolumn{2}{c|}{\textit{Attraction}} & \multicolumn{2}{c|}{\textit{Restaurant}} & \multicolumn{2}{c}{\textit{Taxi}} \\ \hline
\multicolumn{2}{c|}{Training 1\% New Domain} & 19.53 & 77.33 & 44.24 & 85.66 & \textbf{35.88} & \textbf{68.60} & 32.72 & 82.39 & 60.38 & 72.82 \\ \hline
\multicolumn{1}{c|}{\multirow{3}{*}{\begin{tabular}[c]{@{}c@{}}Fine-tuning BM\\ on 1\% new domain\end{tabular}}} & \textit{Naive} & 19.13 & 75.22 & \textbf{59.83} & \textbf{90.63} & 29.39 & 60.73 & \textbf{42.42} & \textbf{86.82} & \textbf{63.81} & \textbf{79.81} \\
\multicolumn{1}{c|}{} & \textit{EWC} & 19.35 & 76.25 & 58.10 & 90.33 & 32.28 & 62.43 & 40.93 & 85.80 & 63.61 & 79.65 \\
\multicolumn{1}{c|}{} & \textit{GEM} & \textbf{19.73} & \textbf{77.92} & 54.31 & 89.55 & 34.73 & 64.37 & 39.24 & 86.05 & 63.16 & 79.27 \\ \hline
\end{tabular}
}
\caption{We run domain expansion experiments by excluding one domain and fine-tuning on that domain. The first row is the base model trained on the four domains. The second row is the results on the four domains after fine-tuning on 1\% new domain data using three different strategies. One can find out that GEM outperforms Naive and EWC fine-tuning in terms of catastrophic forgetting on the four domains. Then, we evaluate the results on new domain for two cases: training from scratch and fine-tuning from the base model. Results show that fine-tuning from the base model usually achieves better results on the new domain compared to training from scratch.}
\label{1_alldomain}
\end{table*}

We visualize the cosine similarity matrix for all possible slot embeddings in Fig.~\ref{slot-viz}. 
Most of the slot embeddings are not close to each other, which is expected because the model only depends on these features as start-of-sentence embeddings to distinguish different slots.
Note that some slots are relatively close because either the values they track may share similar semantic meanings or the slots are correlated. 
For example,~\textit{destination} and~\textit{departure} track names of cities, while ~\textit{people} and~\textit{stay} track numbers. 
On the other hand,~\textit{price range} and \textit{star} in hotel domain are correlated because high-star hotels are usually expensive.

\paragraph{Zero-shot}
We run zero-shot experiments by excluding one domain from the training set.
As shown in Table~\ref{ZeroShot-TABLE}, the \textit{taxi} domain achieves the highest zero-shot performance, 60.58\% on joint goal accuracy, which is close to the result achieved by training on all the \textit{taxi} domain data (76.13\%).
Although performances on the other zero-shot domains are not especially promising, they still achieve around 50 to 65\% slot accuracy without using any in-domain samples. 
The reason why the zero-shot performance on the \textit{taxi} domain is high is because all four slots share similar values with the corresponding slots in the~\textit{train} domain. 

\begin{table}[t!]
\begin{center}
\resizebox{0.8\linewidth}{!}{
\begin{tabular}{r|cc|cc}
\hline
\multicolumn{1}{l|}{\multirow{2}{*}{}} & \multicolumn{2}{c|}{\textbf{Trained Single}} & \multicolumn{2}{c}{\textbf{Zero-Shot}} \\ \cline{2-5} 
\multicolumn{1}{c|}{} & 
\multicolumn{1}{l}{\textit{Joint}} & 
\multicolumn{1}{c|}{\textit{Slot}} & 
\multicolumn{1}{c}{\textit{Joint}} & 
\multicolumn{1}{c}{\textit{Slot}} \\ \hline
\textit{Hotel} & 55.52 & 92.66 & 13.70 & 65.32 \\ \hline
\textit{Train} & 77.71 & 95.30 & 22.37 & 49.31 \\ \hline
\textit{Attraction} & 71.64 & 88.97 & 19.87 & 55.53 \\ \hline
\textit{Restaurant} & 65.35 & 93.28 & 11.52 & 53.43 \\ \hline
\textit{Taxi} & 76.13 & 89.53 & \textbf{60.58} & 73.92 \\ \hline
\end{tabular}
}
\end{center}
\caption{Zero-shot experiments on an unseen domain. In \textit{taxi} domain, our model achieves 60.58\% joint goal accuracy without training on any samples from \textit{taxi} domain. \textit{Trained Single} column is the results achieved by training on 100\% single-domain data as a reference.}
\label{ZeroShot-TABLE}
\end{table}

\paragraph{Domain Expanding}
In this setting, the TRADE model is pre-trained on four domains and a \textit{held-out} domain is reserved for domain expansion to perform fine-tuning. 
After fine-tuning on the new domain, we evaluate the performance of TRADE on 1) the four pre-trained domains and 2) the new domain. We experiment with different fine-tuning strategies.
The \textit{base model} row in Table~\ref{1_alldomain} indicates the results evaluated on the four domains using their in-domain training data, and the \textit{Training 1\% New Domain} row indicates the results achieved by training from scratch using 1\% of the new domain data.
In general, GEM outperforms naive and EWC fine-tuning in terms of overcoming catastrophic forgetting. We also find that pre-training followed by fine-tuning outperforms training from scratch on the single domain.

Fine-tuning TRADE with GEM maintains higher performance on the original four domains.
Take the \textit{hotel} domain as an example, the performance on the four domains after fine-tuning with GEM only drops from 58.98\% to 53.54\% (-5.44\%) on joint accuracy, whereas naive fine-tuning deteriorates the tracking ability, dropping joint goal accuracy to 36.08\% (-22.9\%).


Expanding TRADE from four domains to a new domain achieves better performance than training from scratch on the new domain. 
This observation underscores the advantages of transfer learning with the proposed TRADE model.
For example, our TRADE model achieves 59.83\% joint accuracy after fine-tuning using only 1\% of \textit{Train} domain data, outperforming the training \textit{Train} domain from scratch, which achieves 44.24\% using the same amount of new-domain data.

Finally, when considering \textit{hotel} and \textit{attraction} as new domain, fine-tuning with GEM outperforms the naive fine-tuning approach on the new domain.
To elaborate, GEM obtains 34.73\% joint accuracy on the \textit{attraction} domain, but naive fine-tuning on that domain can only achieve 29.39\%.
This implies that in some cases learning to keep the tracking ability (learned parameters) of the learned domains helps to achieve better performance for the new domain.


\section{Error Analysis}

\begin{figure}[t]
\centering
\includegraphics[width=\linewidth]{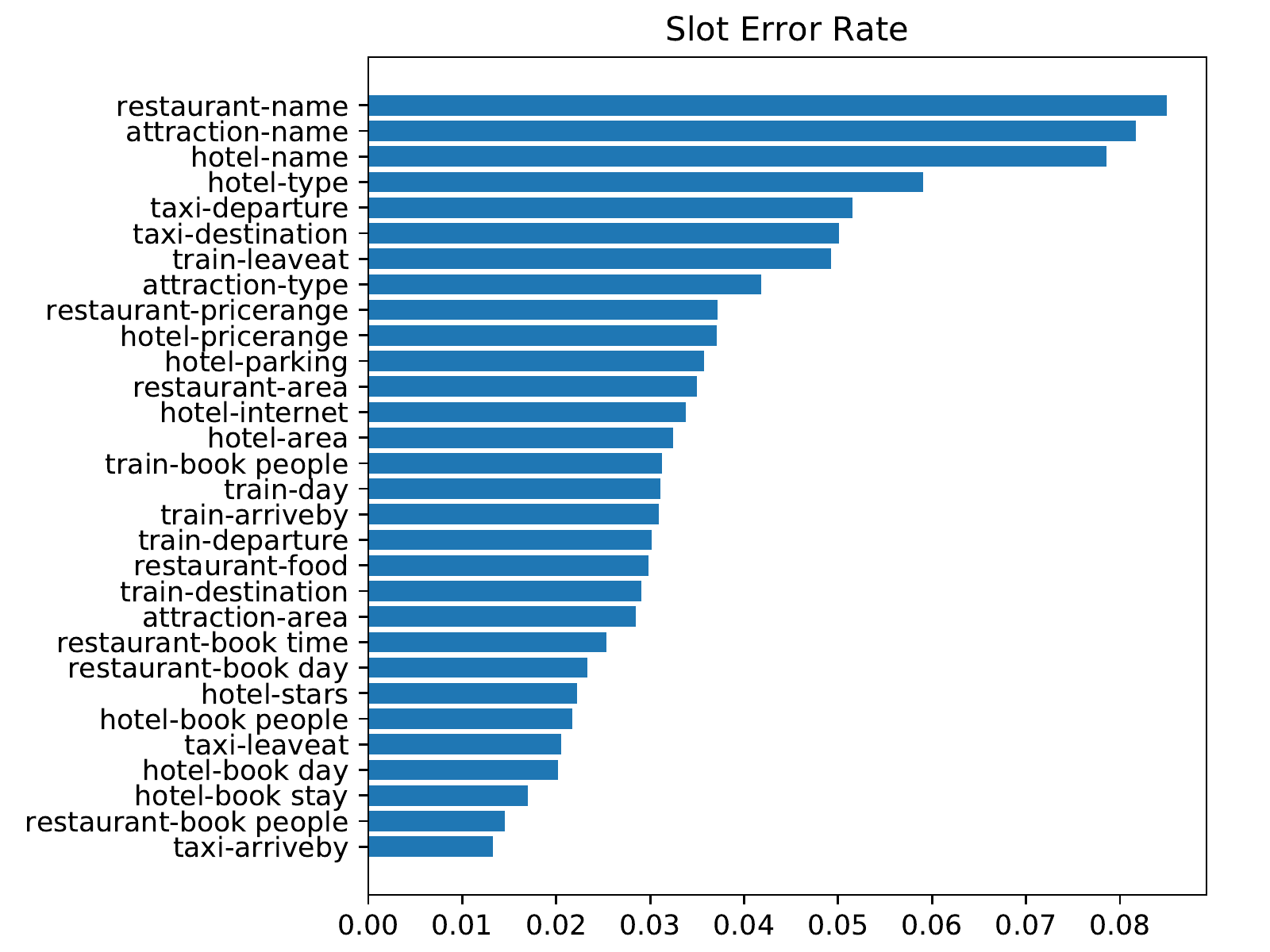}
\caption{Slots error rate on test set of multi-domain training. The \textit{name} slot in \textit{restaurant} domain has the highest error rate, 8.50\%, and the \textit{arrive\_by} slot in \textit{taxi} domain has the lowest error rate, 1.33\%}
\label{error_rate}
\end{figure}

An error analysis of multi-domain training is shown in Fig.~\ref{error_rate}.
Not surprisingly, \textit{name} slots in the \textit{restaurant}, \textit{attraction}, and \textit{hotel} domains have the highest error rates, 8.50\%, 8.17\%, and 7.86\%, respectively.
It is because this slot usually has a large number of possible values that is hard to recognize. 
On the other hand, number-related slots such as \textit{arrive\_by}, \textit{people}, and \textit{stay} usually have the lowest error rates.
We also find that the \textit{type} slot of \textit{hotel} domain has a high error rate, even if it is an easy task with only two possible values in the ontology.
The reason is that labels of the \textit{(hotel, type)} pair are usually missing in the dataset, which makes our prediction incorrect even if it is supposed to be predicted.

\begin{figure}[t!]
    \subfloat[Hotel]{\label{sublable1}
        \includegraphics[width=0.48\linewidth]{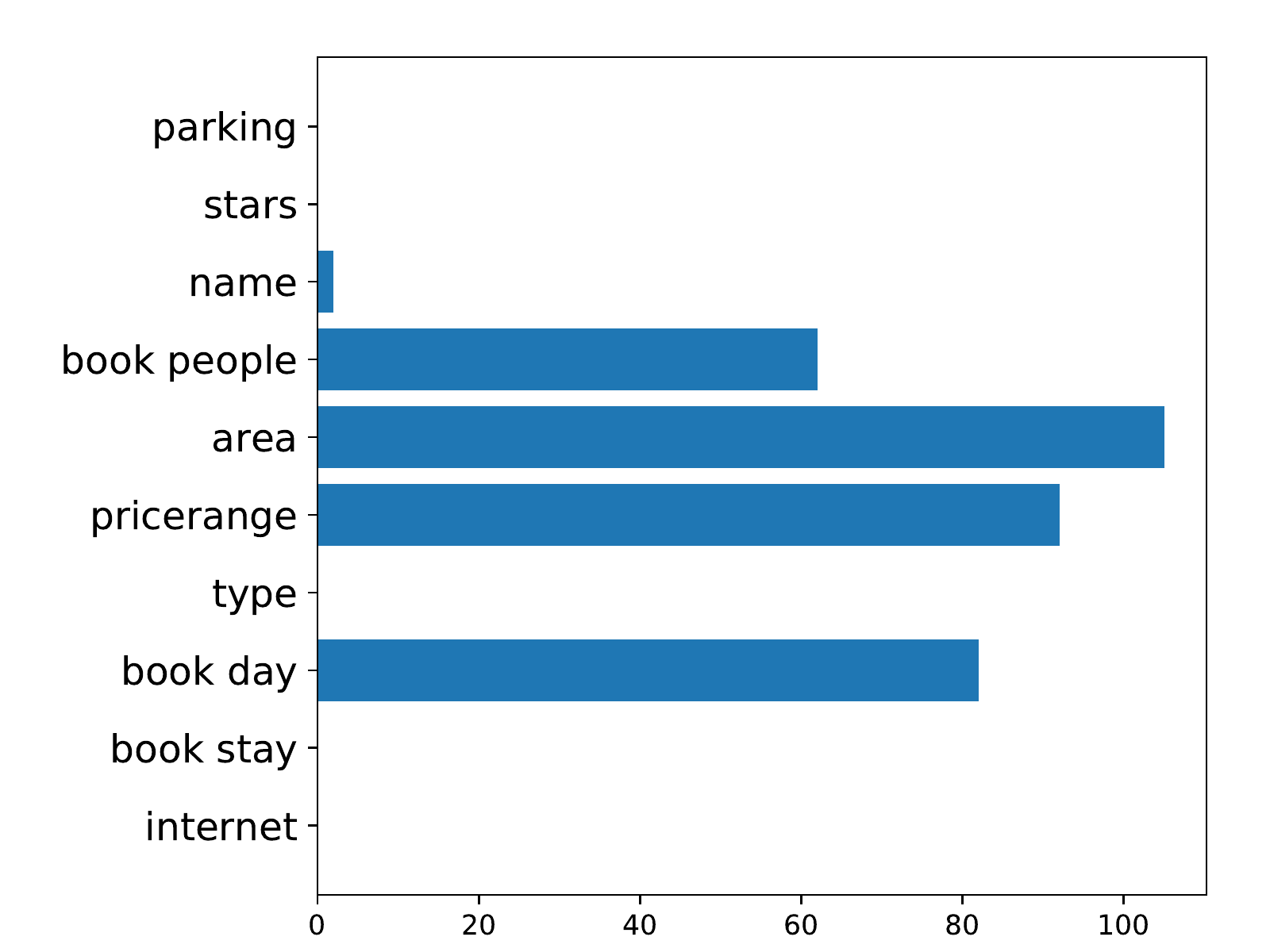}} 
    \hfill
    \subfloat[Restaurant]{\label{sublable2}
        \includegraphics[width=0.48\linewidth]{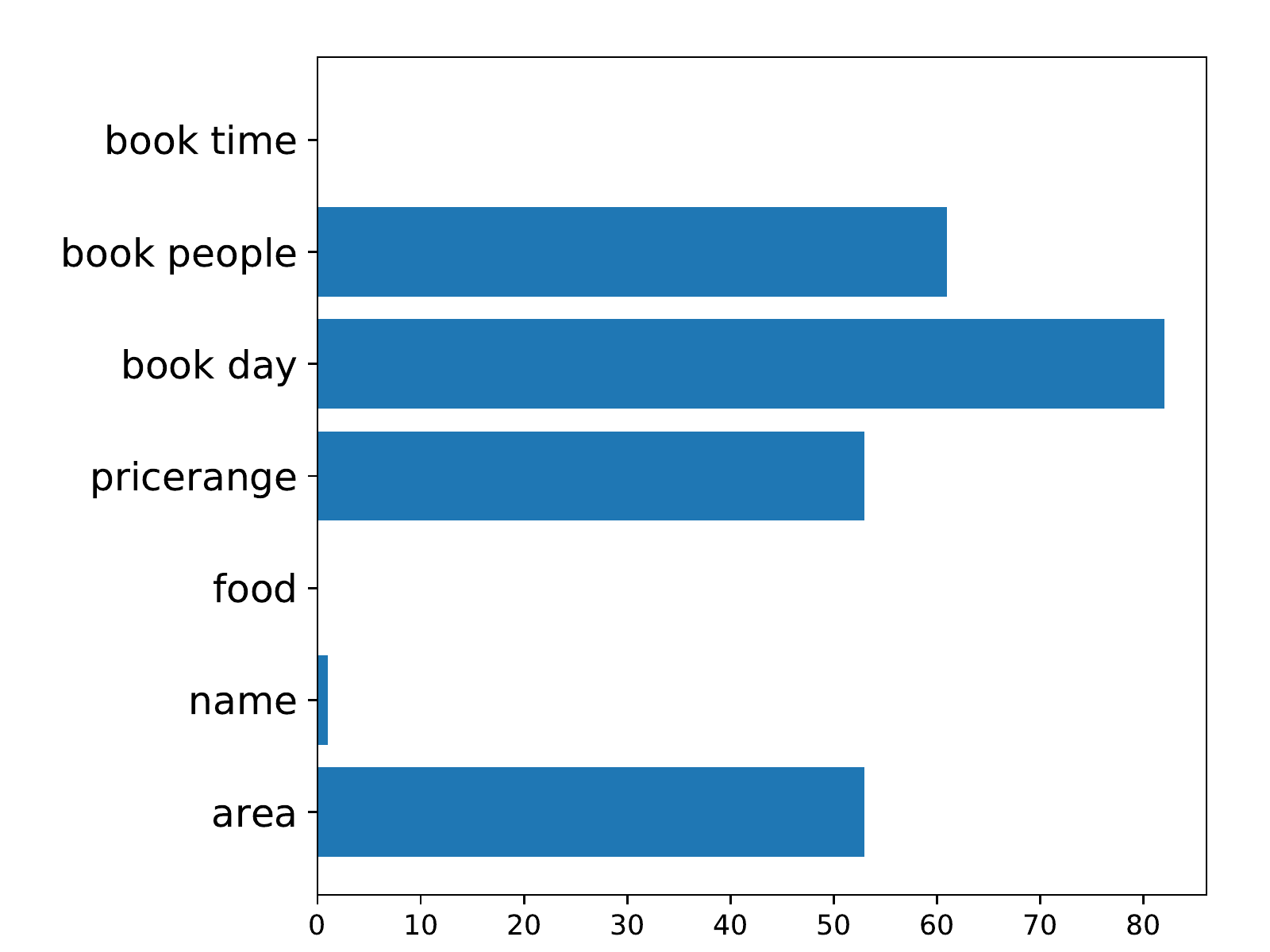}}
    \caption{Zero-shot DST error analysis on (a) \textit{hotel} and (b) restaurant domains. The x-axis represents the correctness number of each slot which has non-empty values. In \textit{hotel} domain, abilities of tracking \textit{people, area, price\_range}, and \textit{day} slots are successfully transferred.}
    \label{fig:zscorrect}
\end{figure}

In Fig.~\ref{fig:zscorrect}, we show the zero-shot analysis of two selected domains, \textit{hotel} and \textit{restaurant}, that have more slots to be tracked. 
To better understand the behavior of knowledge transferring, here we only take labels that are not missing into account, i.e., we ignore data that is labeled as ``none'' because predicting ``none'' is relatively easier for the model.
In both \textit{hotel} and \textit{restaurant} domains, \textit{people, area, price\_range}, and \textit{day} slots are successfully transferred from the other four domains. 
For unseen slots that only appear in one domain, it is very hard for our model to track correctly.
For example, \textit{parking, stars} and \textit{internet} slots are only appeared in \textit{hotel} domain, and the \textit{food} slot is unique to the \textit{restaurant} domain. 

\section{Related Work}
\paragraph{Dialogue State Tracking}
Traditional dialogue state tracking models combine semantics extracted by language understanding modules to estimate the current dialogue states~\cite{williams2007partially,thomson2010bayesian, wang2013simple, williams2014web}, or to jointly learn speech understanding ~\cite{henderson2014word, zilka2015incremental, Wen-WOZ}. One drawback is that they rely on hand-crafted features and complex domain-specific lexicons (besides the ontology), and are difficult to extend and scale to new domains.

~\citet{NBT} use distributional representation learning to leverage semantic information from word embeddings to and resolve lexical/morphological ambiguity.
However, parameters are not shared across slots. 
On the other hand, ~\citet{Nouri2018TowardSN} utilizes global modules to share parameters between slots, and ~\citet{GLAD} uses slot-specific local modules to learn slot features, which has proved to successfully improve tracking of rare slot values. 
~\citet{lei2018sequicity} use a Seq2Seq model to generate belief spans and the delexicalized response at the same time.
~\citet{ren2018towards} propose StateNet that generates a dialogue history representation and compares the distances between this representation and value vectors in the candidate set.
~\citet{P18-1134PtrNet} use the index-based pointer network for different slots, and show the ability to point to unknown values. 
However, many of them require a predefined domain ontology, and the models were only evaluated on single-domain setting (DSTC2).

For multi-domain DST, \citet{rastogi2017scalable} propose a multi-domain approach using two-layer bi-GRU. Although it does not need an ad-hoc state update mechanism, it relies on delexicalization to extract the features. 
~\citet{MDBT} propose a model to jointly track domain and the dialogue states using multiple bi-LSTM. They utilize semantic similarity between utterances and the ontology terms and allow the information to be shared across domains. For a more general overview, readers may refer to the neural dialogue review paper from ~\citet{gao2018neural}. 

\paragraph{Zero/Few-Shot and Continual Learning}
Different components of dialogue systems have previously been used for zero-shot application, e.g., intention classifiers~\cite{chen2016zero}, slot-filling~\cite{bapna2017towards}, and dialogue policy~\cite{gavsic2014gaussian}. 
For language generation,~\citet{Q17-1024} propose single encoder-decoder models for zero-shot machine translation, and ~\citet{zhao2018zero} propose cross-domain zero-shot dialogue generation using action matching.
Moreover, few-shot learning in natural language applications has been applied in semantic parsing~\cite{N18-2115}, machine translation ~\cite{D18-1398}, and text classification~\cite{N18-1109} with meta-learning approaches~\cite{schmidhuber:1987:srl,finn2017model}.
These tasks usually have multiple tasks to perform fast adaptation, instead in our case the number of existing domains are limited.
Lastly, several approaches have been proposed for continual learning in the machine learning community 
~\cite{kirkpatrick2017overcoming,lopez2017gradient,rusu2016progressive,fernando2017pathnet,lee2017overcoming}, especially in image recognition tasks~\cite{aljundi2017expert,rannen2017encoder}. 
The applications within NLP has been comparatively limited, e.g., ~\citet{shu2016lifelong, P17-2023} for opinion mining, ~\citet{D17-1314} for document classification, and ~\citet{lee2017toward} for hybrid code networks~\cite{P17-1062}.

\section{Conclusion}
We introduce a transferable dialogue state generator for multi-domain dialogue state tracking, which learns to track states without any predefined domain ontology.
TRADE shares all of its parameters across multiple domains and achieves state-of-the-art joint goal accuracy and slot accuracy on the MultiWOZ dataset for five different domains.
Moreover, domain sharing enables TRADE to perform zero-shot DST for unseen domains and to quickly adapt to few-shot domains without forgetting the learned ones.
In future work, transferring knowledge from other resources can be applied to further improve zero-shot performance, and collecting a dataset with a large number of domains is able to facilitate the application and study of meta-learning techniques within multi-domain DST.

\section*{Acknowledgments}
This work is partially funded by MRP/055/18 of the Innovation Technology Commission, of the Hong Kong University of Science and Technology (HKUST).

\bibliography{acl2019}
\bibliographystyle{acl_natbib}

\end{document}